\newcommand{\ignore}[1]{}

\newcommand{\nscomment}[1]{\footnote{!!NS!! #1}}
\newcommand{\hmcomment}[1]{\footnote{!!HM!! #1}}

\renewcommand{\nscomment}[1]{}
\renewcommand{\hmcomment}[1]{}

\documentclass[
  ,final            
]
{aipproc}

\layoutstyle{6x9}

\begin{document}

\title{Understanding physics from interconnected data}


\author{Nikita A. Sakhanenko}{
  address={E548, P-22, Los Alamos National Laboratory}
  ,altaddress={CS Department, University of New Mexico}
}

\author{Hanna E. Makaruk}{
  address={E548, P-22, Los Alamos National Laboratory}
}

\begin{abstract}

Metal melting on release after explosion is a physical system far from equilibrium. 
A complete physical model of this system does not exist, because many interrelated 
effects have to be considered. General methodology needs to be developed so as 
to describe and understand physical phenomena involved. 

The high noise of the data, moving blur of images, the high degree of uncertainty 
due to the different types of sensors, and the information entangled 
and hidden inside the noisy images makes reasoning about the physical processes 
very difficult. Major  problems include proper information 
extraction and the problem of reconstruction, as well as prediction of 
the missing data. In this paper, several techniques addressing the 
first problem are given, building the basis for tackling the second 
problem.

\end{abstract}

\maketitle


\section{Introduction}

Metal melting on release after explosion is considered here as 
an example of a physical system far from equilibrium. The goal is 
not only to describe and understand this system, but also to develop 
a more general methodology to approach similar problems. Due to 
the highly non-equilibrium nature of the 
process and the simultaneous interconnection of many different physical 
effects, a complete physical model of the process does not exist.
Bayesian and correlational analysis of the 
data, as well as other data analysis methods, 
is a step in the direction of building the physical model.
The ultimate task of this work is to find and utilize all 
possible dependencies within the system.

Metal melting on release possesses a number of different characteristics, 
some of which seem to be incompatible with each other. These include time, 
metal type, thickness of the material, etc. One can see that the problem of 
finding the physical model of the system is connected to the problem of 
combining its measurable properties in the case of minimal compatibility. 
Moreover, some of these characteristics, such as time and thickness, belong 
to a potentially infinite domain. On the other hand, being able to consider  
all the features of the system together is very important in understanding the 
underlying physical model.

Another important problem is the proper extraction of information from the  raw 
experimental data. The extreme conditions of the environment restrict the choice
of the sensors to use. Particularly, images are generated by proton radiography (PRAD) 
method. To achieve proper contrast, protons for each image are collected 
for a period of time that is relatively long, in comparison to the time scale of the physical phenomena, 
which in turn creates blurry, noisy, relatively low 
contrast proton-radiographic images (by typical standards of image processing). These 
images are a ``gold mine'' of the information essential to a specific 
goal; hence, feature extraction from the images is driven by the goal. 
On the other hand, they are significantly more difficult to process then usual images. 
Parameters like velocity of different parts of the system obtained from the 
images constitute the system representation. 
Standard image processing approaches, like gradient-based 
contour detection, however, can not be applied due to insufficient quality of the 
images. Other methods of contour detection and velocity calculations 
developed in the project are presented.

\section{Experiment}

\begin{figure}
  \includegraphics[height=.3\textheight]{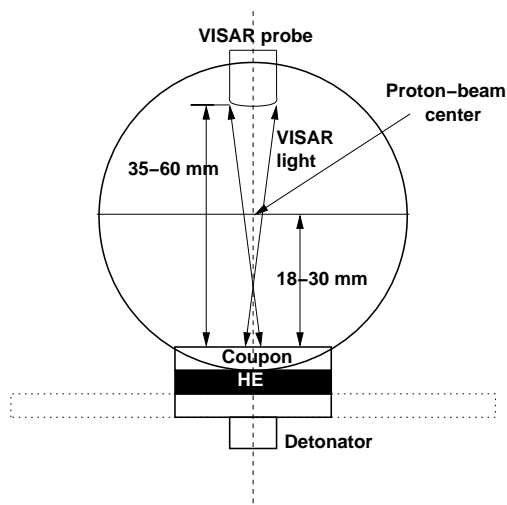}
  \caption{Schematic representation of the experiment configuration.}
\label{fig:scheme}
\end{figure}

The configuration of the experiment is given in figure \ref{fig:scheme}.
As seen in the figure, a metal coupon is connected to a high explosive (HE)
cylinder 2-in. in diameter and 0.5-in. thick that is point initiated with
a detonator centered on the charge. \ignore{There are several diagnostic tools used
during the experiment, such as VISAR that captures the point velocity by
means of a laser and Proton Radiography that produces images of the experimental
system at a single time unit.} A VISAR probe is positioned 35-60 mm from
the metal surface, whereas the center of the proton beam is 18-30 mm from
the surface. The experimental data for this work were obtained from 
the research of D. B. Holtkamp et al \cite{Holtkamp:2003}.

All considered experiments were performed under the same conditions, 
including the same type of high explosives with the point ignition positioned
in the center, and the same diameter of metal coupons. \ignore{ On the other hand,
thickness of the coupon and the type of metal were varied during the
experimentation.} In this paper, the work on tin samples of different 
thicknesses and their comparison is presented. Investigation of experiments
with other types of metal and, possibly, with other types of ignition is
the next step of this project.

\subsection{Types of measurements}

This work is based on two kinds of experimental data: 
numerical time series of velocities measured 
by VISAR that utilizes the Doppler effect of a laser beam \cite{Hemsing:1983}, and 
time series of up to 20 images 
per experiment obtained using Proton Radiography (PRAD) \cite{Hogan:1999}. 

\begin{figure}
  \includegraphics[height=.35\textheight]{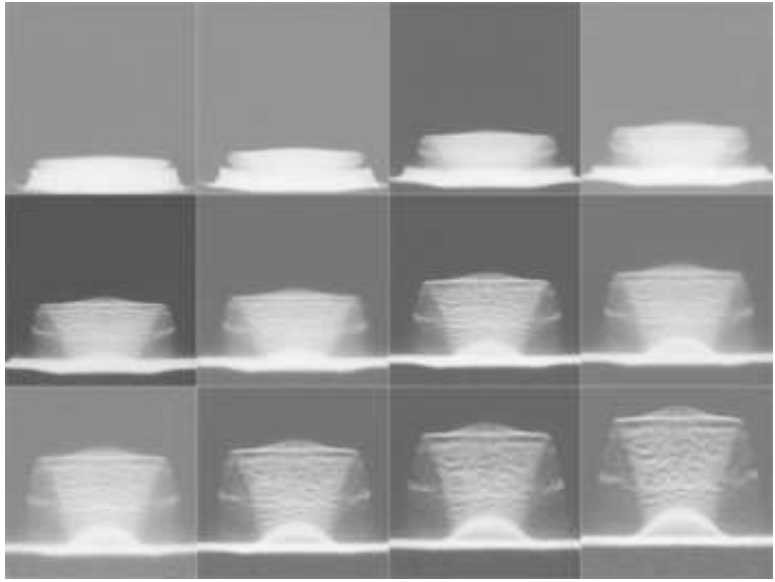}
  \caption{Structure evolution of the melting metal sample after explosion
of 0.4375-in. thick coupon.}
\label{fig:evol}
\end{figure}

In figure \ref{fig:evol}, a series of PRAD images illustrating the melting during 
explosion (up to 57 usec ) of a 0.4375-in. thick
coupon. The first image shows the coupon immediately after 
the ignition, while 
next images made approximately each 3 usec  show the 
consecutive stages of the process.  The data for each image were collected for 3.28 usec. 
Time series of images are very helpful
in identifying different phases of the process. Note that a PRAD image is scaled in 
physical coordinates such that 1 pixel is equivalent to 0.01 mm$^2$.

\begin{figure}
  \includegraphics[height=.3\textheight]{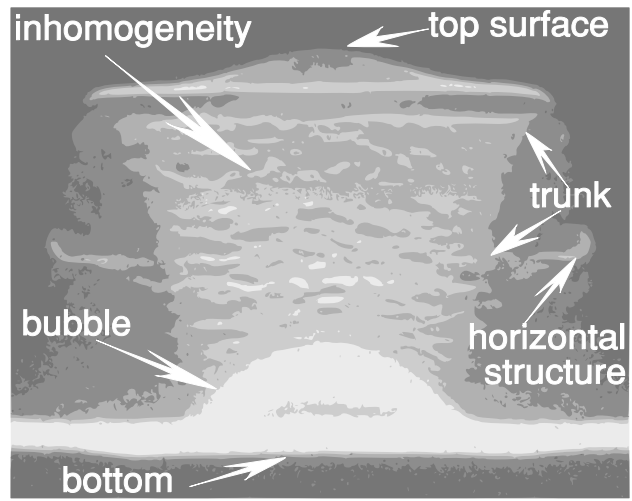}
  \caption{Main structures observed on a PRAD image. The image is captured 
about 50 usec after an ignition of 0.4375-in. thick tin coupon.}
\label{fig:struct}
\end{figure}

In figure \ref{fig:struct}, one can identify structures from the PRAD image of a melting 
coupon. The {\em top surface} could be solid in some experiments,\nscomment{Later on
we should tell what are the possibilities, based on our observations} 
while the {\em bubble}, 
together with the {\em bottom}, containing the majority of material,\nscomment{Later
we should show that the bubble indeed contain most of material} are 
always completely melted.\nscomment{This is a very strong statement, which should be
addressed later: spherical shape, formation of the void inside... BTW,
why it is void should be explained too} In some experiments, the {\em horizontal 
structure} and {\em inhomogeneity} in the {\em trunk} area may be not present\nscomment{Later
we need to state more clear that it is not just invisible, but not present, and why}. 
The observed structure has axial symmetry due to the point ignition of the explosives, 
which is also responsible for the sheer waves observed in the material.

\begin{figure}
  \includegraphics[height=.4\textheight]{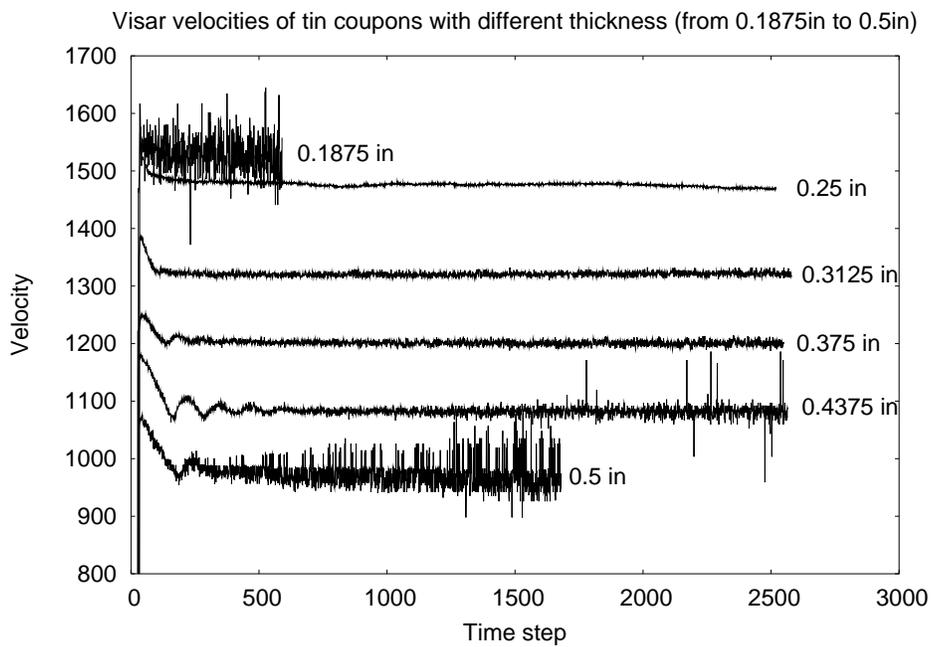}
  \caption{Time series of velocities measured by VISAR for tin coupons. 
The six series showed on the figure correspond to six different thicknesses in increasing order 
from top to bottom, ranging between 0.1875 and 0.5 inches.}
\label{fig:visar}
\end{figure}

Using VISAR, one can measure a velocity of the central point on the top surface.
In figure \ref{fig:visar}, time series of such velocities corresponding to 
various thicknesses of the tin coupon are showed. As can be observed in 
figure \ref{fig:visar}, the thinnest coupon melted completely almost immediately after explosion, 
hence the VISAR reading came out very noisy (see the very top time series of the figure). 
Many observations can be linked to the increasing thickness of the coupon. For example,
as the thickness increases, the average velocity decreases, the noise of the time series
increases, the magnitude of the fluctuations increases, and the subsequent time before these
fluctuations increases.
\ignore{
The average velocity of each time series decreases
with increasing thickness of the coupon. Another observation one can make is that the
noise of a time series increases as the coupon thickness goes up. In addition,
it is possible to identify an oscillation in a single time series. This fluctuation
becomes more apparent during the experiments on thick coupons. Finally, one can 
see from figure \ref{fig:visar} that the time before the first oscillation occurs increases
as the coupon in the experiment becomes larger.\nscomment{All the observations here should
be discussed and explained later on.}
}

\section{Problem}

One of the main goals of this project is to quantify all possible information about 
the physical system underlying the experiment. At first, appropriate parameters
should be identified and the relationships among them studied. These parameters
include time and sample thickness. It is important to construct
a data manifold \cite{Amari:2000} in multidimensional data space that contains all the
measured information at a single moment of time. \ignore{Note that the qualitative description
of the process is required for successful manifold construction.} Analysis of this data
manifold evolution allows for the predictability of the system behavior on a longer 
time scale. Particularly, the missing data of one of the manifold's dimensions (corresponding
to one of the sensors) can be estimated via other fully specified data sets of 
the manifold. In a sense, a data manifold reflects the data and qualitative
description of the system, and thus provides {\em history} in Bayesian reasoning.

It is important to make minimal theoretical assumptions while discovering the origins
of the processes of the system and predicting their future behavior. In reality,
assumptions about a model are unavoidable. However, tracing their influence on the
results and understanding is crucial.

There are many different features of the given data that make the reasoning
about the physical system difficult. These include gaps and uncertainty in the data
due to the stochastic nature of the system. The given data is highly
uneven, overlapped, and multidimensional. For instance, the VISAR data spans
the initial period of time very densely, however at each time step it 
provides only a small amount of information. PRAD images, on the other hand, 
cover a longer period of time than VISAR data very sparsely, although, at each
covered time step, a PRAD image provides a huge amount of information about 
many different parameters of the system. Data for one image is collected for 
a relatively long time (about 3.4 usec), taking into account that the speed 
of ejected metal is frequently of the order of 1km/s, while the time/space scale of 
physically interesting details is rather small. 

\ignore{Even though PRAD images contain a lot of information, it is all entangled
and buried inside of each image making detailed image processing essential.}
Due to this large quantity of information in these images, image processing is
essential.
In addition to the difficulties of extraction of some information, even with 
cutting edge techniques, image processing is complicated because of specific
features of PRAD imagery, including noise and granularity of the image, image blur, 
under- and over-shooting effects on the borders between different densities
areas, scattering, and the fact that the axial symmetry assumption of PRAD makes
non-cylindrical objects difficult to track.

Since the process being investigated is far from equilibrium, there is no
direct way to measure thermodynamic quantities. These data can be obtained
only under the assumption of an underlying numerical model, 
influencing the results and raise non-trivial questions about their credibility.

\section{Methodology}

The simplest way of analyzing different types of data describing one
process is to work with each data type independently, due to the difficulty 
comparing essentially distinct sorts of data with significant error 
variability. However, it is strongly believed by the authors that besides considering
parts of data, the data set must be analyzed as a whole in order to gain
more knowledge of the process, especially in the presence of expensive,
hard-to-repeat experiments. In particular, in order to solve 
the problem of data gaps, interpolation
can be used. However, depending on the size of a gap, this approach might
not produce the best solution. A higher-order analysis that accounts for the
data from other sensors can be performed so as to capture the behavior
of the process in the gap.

\subsection{Image analysis}

The goal of the first stage of the project is to gather information from
the available data, a large portion of which is represented with images.
Hence, image processing driven by the features of an image dictates further
steps of the project.

In order to improve the performance of image analysis tools, the images are
denoised. Since different parts of the same image require different amount
of denoising, an additional heat-based denoising is performed.

\begin{figure}
  \includegraphics[height=.095\textheight]{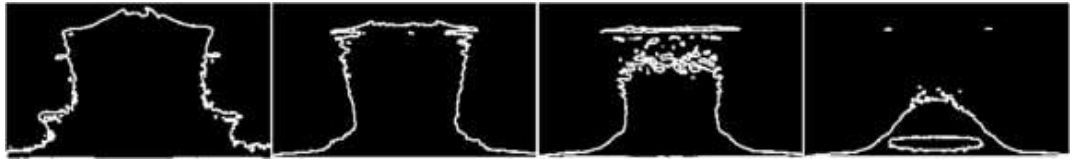}
  \caption{Different contours detected from a single PRAD image
using 1-bit erosion method with the erosion threshold increasing 
from left to right.}
\label{fig:contour}
\end{figure}

\begin{figure}
  \includegraphics[height=.2\textheight]{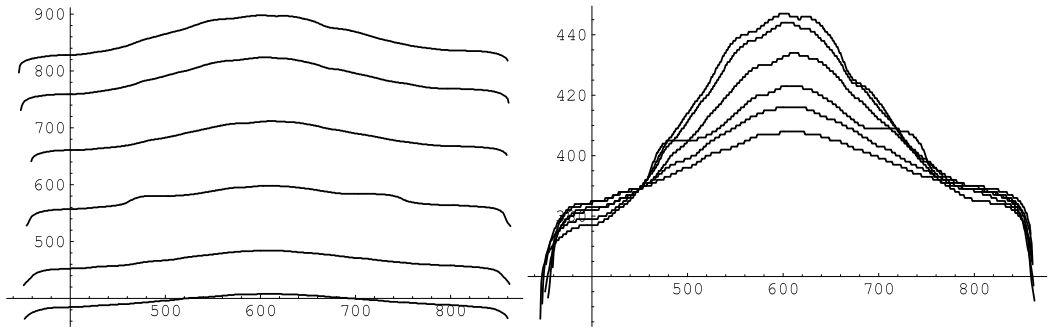}
  \caption{Evolution of the top surface: the movement of the top surface 
from the bottom to the top in
physical coordinates (left); the overlapped surfaces emphasizing the shape 
evolution (right). The axes correspond to physical coordinates.
It can be seen that the central part of the top surface is moving slightly faster than the 
sides, which, however, does not tell whether the surface is melted or not.}
\label{fig:top}
\end{figure}

\begin{figure}
  \includegraphics[height=.2\textheight]{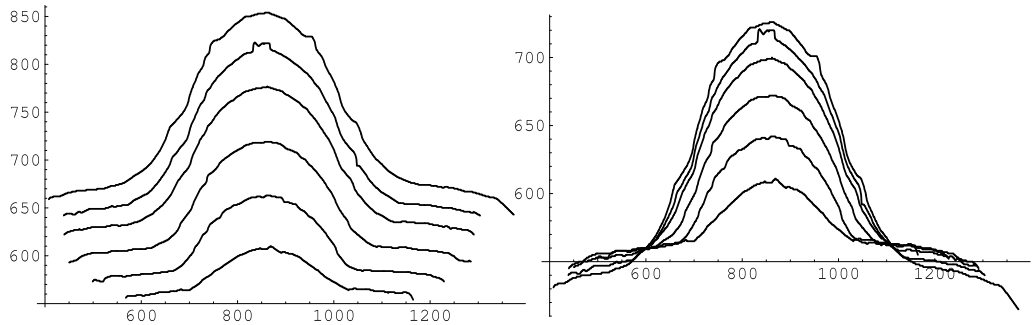}
  \caption{Evolution of the bubble: the movement of the bubble from
the bottom to the top in
physical coordinates (left); the overlapped bubble surfaces
emphasizing the shape evolution (right). It can be seen that 
bubble's curvature is preserved, reflecting the shape of the initial shock wave.}
\label{fig:bubble}
\end{figure}

Due to low contrast and high granularity of PRAD images, gradient-based
contour detection methods perform poorly. As an alternative, the following
method, called {\em 1-bit erosion}, is proposed. The gray scale image
is converted into a 1-bit approximation by choosing a specific threshold.
After eroding the black area one pixel deep and subtracting it from the
1-bit image, a continuous contour emerges (see figure \ref{fig:contour}).
The 1-bit erosion method, specially designed to extract contours from blurred images, 
is fast and robust. As can be seen in figure \ref{fig:contour},
one can catch contours of different structures of an image by choosing
different erosion thresholds. However, the method is sensitive to the contrast
of images: a threshold corresponding to a particular contour on one image 
corresponds to a different contour on the other image, and thus has to be
adjusted manually. After performing contour detection on PRAD images, time series of contours
are collected and analyzed (see figures \ref{fig:top} and \ref{fig:bubble}).


\begin{figure}
  \hfill
  \begin{minipage}[t]{.45\textwidth}
    \begin{center}  
      \includegraphics[height=.21\textheight]{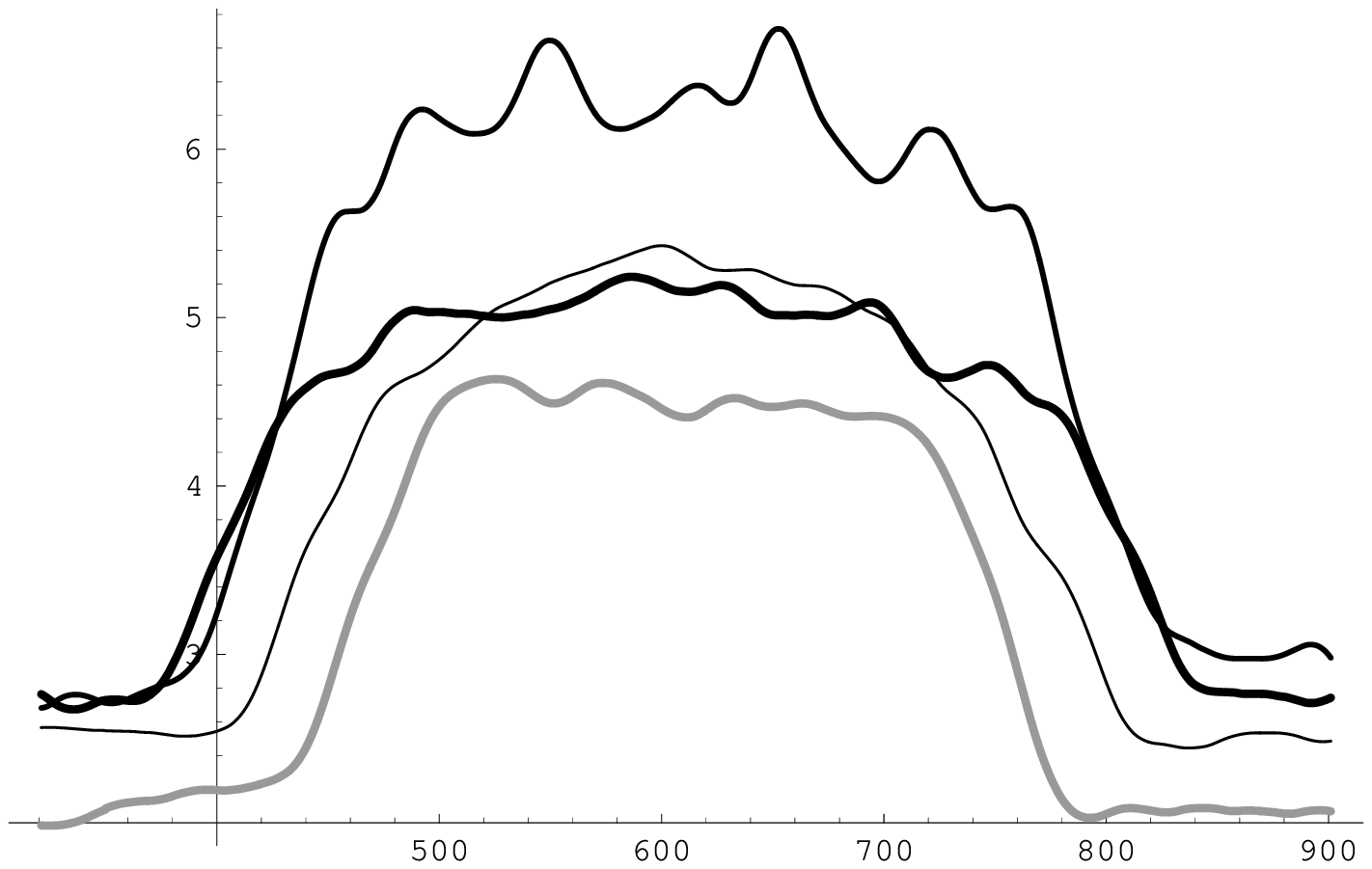}
    \end{center}
  \end{minipage}
  \hfill
  \begin{minipage}[t]{.45\textwidth}
    \begin{center}  
      \includegraphics[height=.21\textheight]{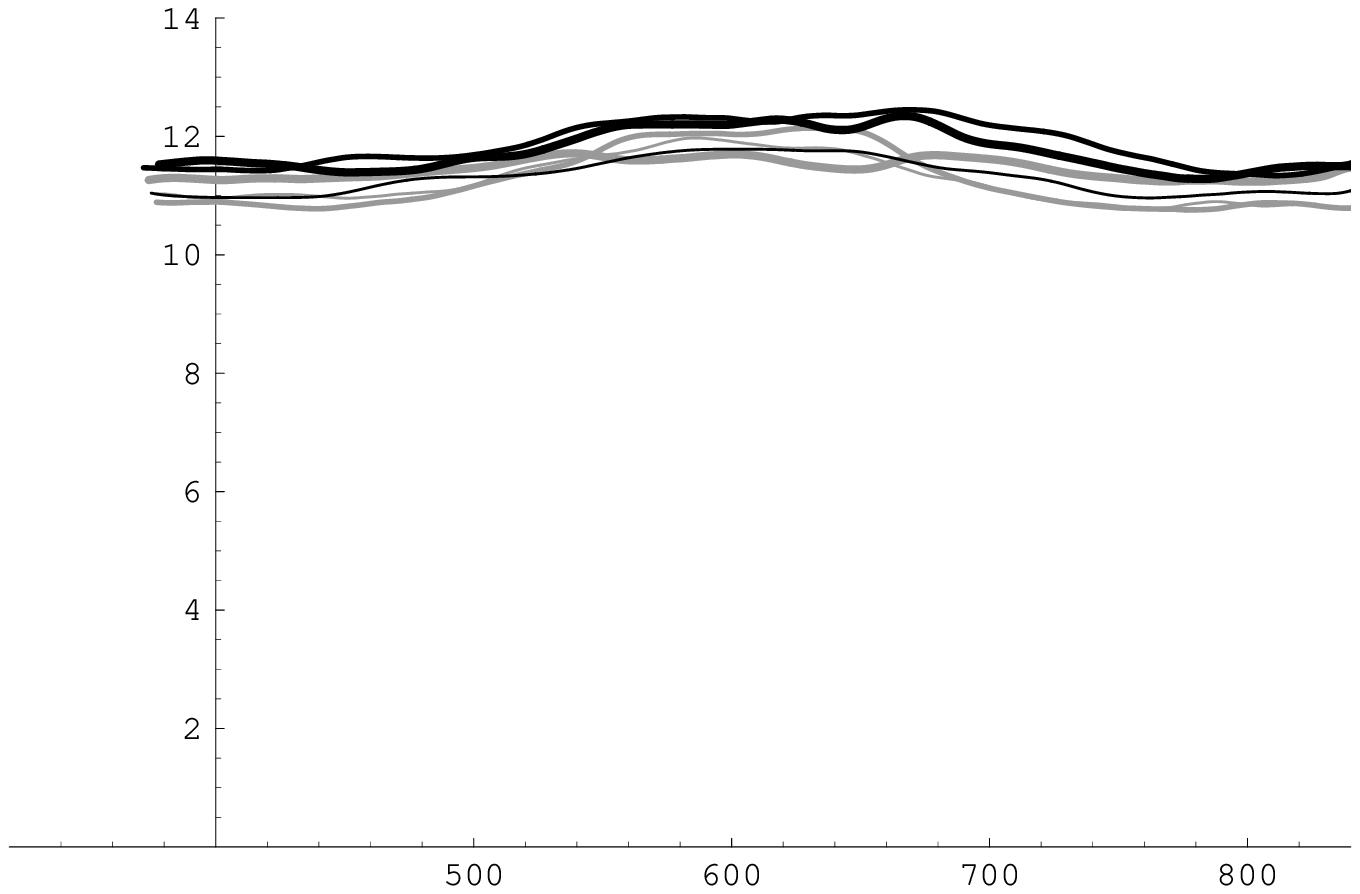}
    \end{center}
  \end{minipage}
  \hfill
  \caption{Changes in velocities of different parts of the system:
velocities of the bubble (left); velocities of the top surface (right).
For readability only five velocities are given. The time progresses from
the gray line to the black one and from the thin one to the thick. The 
$x$-axis shows physical coordinates and the $y$-axis shows velocity.}
\label{fig:vel}
\end{figure}
%
Using the space and time coordinates, velocities of various parts of the
system are calculated. In figure \ref{fig:vel}, one can see the velocities 
of the bubble and the top surface. As expected, the top surface moves faster
than other parts of the system, while its center part evolves even faster.
On the other hand, a semi-spherical shock wave approaches the center part 
of the building bubble first, and the bubble, as the wave progresses, grows
up involving more and more area around itself.

\section{Discussion}

A complicated system composed of multiple processes, for which a mathematical
model is not known, is frequently described by a set of various types of data
measured by different sensors. In this paper, the experiment with melting of
a tin coupon with different thickness is considered, making a first step towards
the ultimate goal of understanding this physical system, which is equally important
as developing a general methodology for analyzing and modeling its data.

So far the preliminary image analysis was performed on PRAD images, revealing
the shape evolution of the system's structures and their velocity fields.
Consequently, the numerical time series (as opposed to series of images) are 
obtained that allow their comparison to the data from other sensors, such as
VISAR. 

There are two major future directions of the project that are closely related 
to each other: physical and numerical. The former concerns with the physical
origins and phenomena of the system and creation of its physical model, while
the later covers the numerical changes in the system and predictions of its 
further evolution. Potentially, there are several machine learning approaches
to the numerical part of the project including Hidden Markov Models (HMMs) and
Kalman Filters, Markov Networks and Ising Models (or even Cellular Automata).
In particular, it seems interesting to attempt building and training HMM on VISAR
data \cite{Luger:2005} and then comparing HMM's predictions with PRAD data.


\vspace*{-.05\textwidth}
\begin{theacknowledgments}

The authors are thankful to David Holtkamp and Joysree Aubrey for the experimental data and 
numerous long and thought-provoking discussions concerning both the experimental data and 
different ways of analyzing it. Special thanks to Leif Hopkins for 
helping generating the images for this paper.
This work was supported by the Department of Energy under the ADAPT program.
\vspace*{-.05\textwidth}
\end{theacknowledgments}



\bibliographystyle{aipproc}   

\bibliography{maxEnt}

\hyphenation{Post-Script Sprin-ger}
\begin{thebibliography}{5}
\expandafter\ifx\csname natexlab\endcsname\relax\def\natexlab#1{#1}\fi
\providecommand{\enquote}[1]{``#1''}
\expandafter\ifx\csname url\endcsname\relax
  \def\url#1{\texttt{#1}}\fi
\expandafter\ifx\csname urlprefix\endcsname\relax\def\urlprefix{URL }\fi
\providecommand{\eprint}[2][]{\url{#2}}

\bibitem[Holtkamp and et~al.(2003)]{Holtkamp:2003}
D.~B. Holtkamp, and et~al., \enquote{A Survey of High Explosive-Induced Damage
  and Spall in Selected Materials Using Proton Radiography,} in \emph{Shock
  Compression of Condensed Matter-2003}, edited by M.~D. Furnish, and et~al.,
  AIP Conference Proceedings 706, American Institute of Physics, New York,
  2003, pp. 477--482.

\bibitem[Hemsing(1983)]{Hemsing:1983}
W.~F. Hemsing, \enquote{VISAR: Some Things You Should Know,} in \emph{SPIE Vol
  427}, edited by D.~L. Paisley, SPIE, 1983, pp. 144--148.

\bibitem[Hogan and et~al.(1999)]{Hogan:1999}
G.~E. Hogan, and et~al., \enquote{Proton Radiography,} in \emph{Particle
  Accelerator Conference}, edited by A.~Luccio, and W.~MacKa, IEEE, Piscataway,
  1999, vol.~1 of \emph{Particle Accelerator Conference Proceedings}, pp.
  579--583.

\bibitem[Amari and Nagaoka(2000)]{Amari:2000}
S.~Amari, and H.~Nagaoka, \emph{Methods of Information Geometry}, vol. 191 of
  \emph{Translations of Mathematical Monographs}, American Mathematical
  Society, Oxford University Press, 2000, ISBN 0-8218-0531-2.

\bibitem[Chakrabarti et~al.(2005)]{Luger:2005}
C.~Chakrabarti, R.~Rammohan, and G.~F. Luger, \enquote{A First-Order Stochastic
  Modeling Language for Diagnosis,} in \emph{FLAIRS Conference}, edited by
  I.~Russell, and Z.~Markov, FLAIRS Conference Proceedings, AAAI Press, 2005,
  pp. 623--628.

\end{thebibliography}

\IfFileExists{\jobname.bbl}{}
 {\typeout{}
  \typeout{******************************************}
  \typeout{** Please run "bibtex \jobname" to obtain}
  \typeout{** the bibliography and then re-run LaTeX}
  \typeout{** twice to fix the references!}
  \typeout{******************************************}
  \typeout{}
 }

\end{document}